\documentclass[wcp]{jmlr}

\usepackage{longtable}

\usepackage{booktabs}

\usepackage{lineno}

\makeatletter
\let\Ginclude@graphics\@org@Ginclude@graphics 
\makeatother

\jmlrvolume{304,}
\jmlryear{2025}
\jmlrworkshop{ACML 2025}

\title[Direct Quantized Training of Language Models with Stochastic Rounding]{Direct Quantized Training of Language Models with Stochastic Rounding}

  \author{\Name{Kaiyan Zhao}\normalfont{\textsuperscript{1}} \Email{kaiyan1006@logos.t.u-tokyo.ac.jp}\\
  \Name{Tsuguchika Tabaru}\normalfont{\textsuperscript{2}}
  \Email{tabaru@fujitsu.com}\\
  \Name{Kenichi Kobayashi}\normalfont{\textsuperscript{2}} \Email{kenichi@fujitsu.com}\\
  \Name{Takumi Honda}\normalfont{\textsuperscript{2}} \Email{honda.takumi@fujitsu.com}\\
  \Name{Masafumi Yamazaki}\normalfont{\textsuperscript{2}} \Email{m.yamazaki@fujitsu.com}\\
  \Name{Yoshimasa Tsuruoka}\normalfont{\textsuperscript{1}} \Email{tsuruoka@logos.t.u-tokyo.ac.jp}\\
 \normalfont{\textsuperscript{1}}\addr The University of Tokyo, Tokyo, Japan \\
 \normalfont{\textsuperscript{2}}\addr Fujitsu Limited, Kawasaki, Japan
}
\editors{Hung-yi Lee and Tongliang Liu}

\begin{document}

\makeatletter
\let \@jmlrpages \@empty
\makeatother

\maketitle

\begin{abstract}
Although recent quantized Large Language Models, such as BitNet, have paved the way for significant reduction in memory usage during deployment with binary or ternary weights, training these models still demands substantial memory footprints. This is partly because high-precision~(i.e., unquantized) weights required for straight-through estimation must be maintained throughout the whole training process.
To address this, we explore directly updating the quantized low-precision weights without relying on straight-through estimation during backpropagation, aiming to save memory usage during training. Specifically, we employ a stochastic rounding technique to minimize the information loss caused by the use of low-bit weights throughout training. Experimental results on our LLaMA-structured models of various sizes indicate that (1) training with only low-precision weights is feasible even when they are constrained to ternary values; (2) extending the bit width to 8 bits achieves performance on par with BitNet b1.58; (3) our models remain robust to precision scaling and memory reduction, showing minimal performance degradation when moving from FP32 to lower-memory environments~(BF16/FP8); and (4) our models also support inference using ternary weights, showcasing their flexibility in deployment.\footnote{Code is available at \url{https://github.com/KYuuto1006/DQT}.}
\end{abstract}

\section{Introduction}
Large Language Models~(LLMs) have become a promising solution for a wide range of Natural Language Processing~(NLP) tasks, including machine translation~\citep{xu2024a, wu-etal-2024-word, miao-etal-2025-improving}, reasoning~\citep{chatgpt, openai2024gpt4technicalreport, 10.5555/3600270.3601883, miao2024improvingarithmeticreasoningability} and multimodal understanding~\citep{huang2025jointfusionencodingadvancing, mao2025deepresonanceenhancingmultimodalmusic}. However, their development is challenged by the need for vast datasets and substantial computational resources, especially as the size of current LLMs continually grows larger~\citep{duan2024efficienttraininglargelanguage}. 

Quantization, which involves converting high-precision parameter matrices into lower-precision formats, has emerged as an effective approach for enabling resource-efficient LLMs. Traditional quantization methods can be divided into two categories: Post-Training Quantization~(PTQ) and Quantization-Aware Training~(QAT). PTQ reduces the bit precision of weight matrices in an already pre-trained LLM~\citep{NEURIPS2019_c0a62e13, frantar2023gptq}, while QAT incorporates quantization during training, enabling the model to adapt to low-bit precision throughout the learning process~\citep{Jacob_2018_CVPR}. 

\begin{figure*}[t]   
    \centering
    \includegraphics[width=0.8\textwidth]{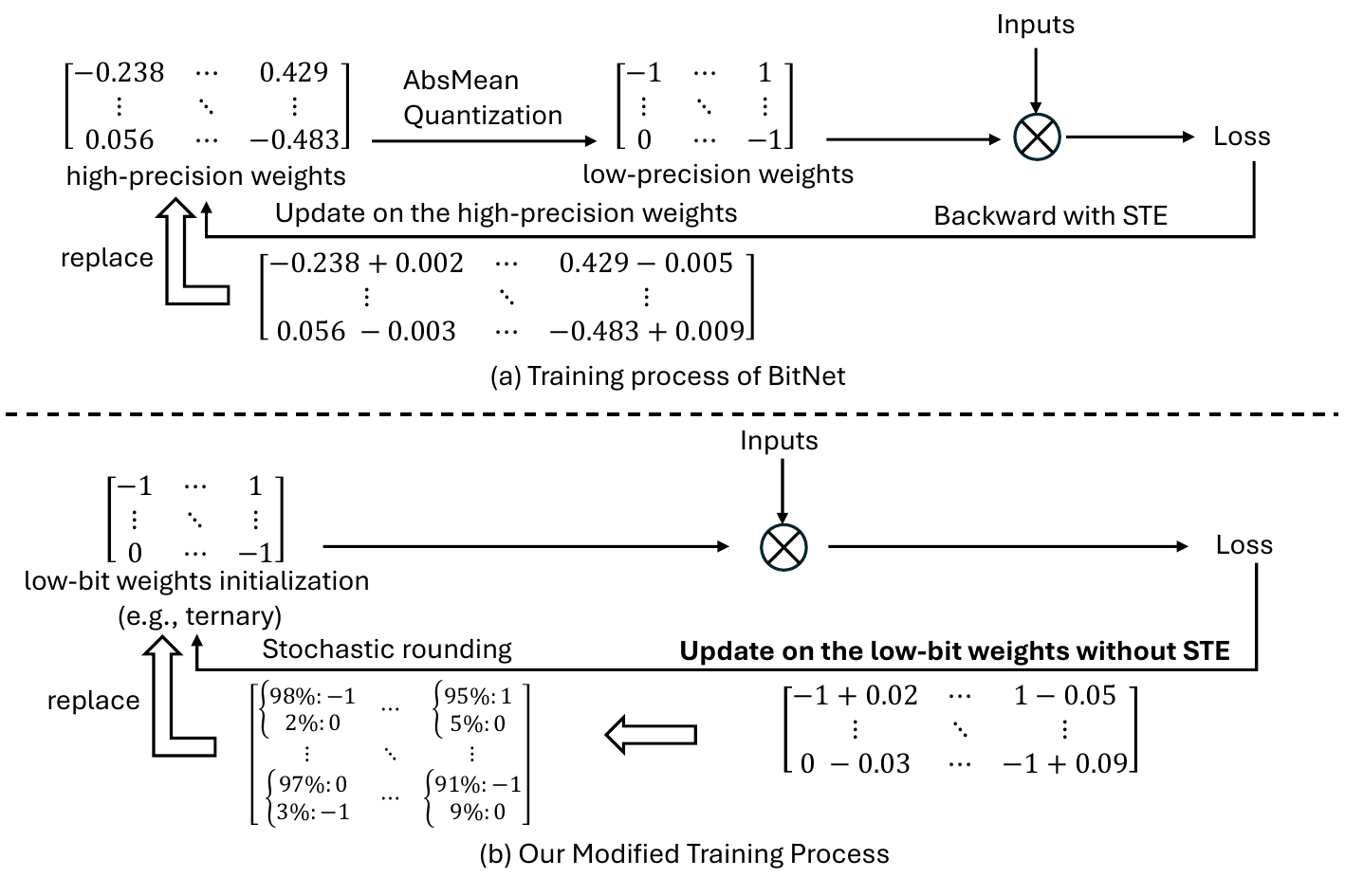}
    \caption{Comparison of the training process for BitNet and our modified one. Upper: The training process for BitNet, where the original high precision weights are updated with the straight-through estimator in backward process. Lower: We directly update the low precision weights with stochastic rounding, eliminating the need to quantize the weight matrices in each training step and keeping weight matrices always at low-bit. We provide an 8-bit example in Supplementary Material, Figure 3.}
    \label{fig_1}

\end{figure*}

Recently, a QAT method, BitNet~\citep{bitnet, b1.58}, has shown the feasibility of quantizing full-precision~(FP32) transformers into binary or ternary models from scratch, while maintaining competitive performance with unquantized ones. We illustrate the training process of traditional QAT methods such as BitNet in Figure~\ref{fig_1} (a). After computing the loss based on the quantized weights and inputs, the loss is backpropagated to update the original high-precision weights via the straight-through estimator~(STE)~\citep{straight_through}. These weights are then re-quantized in each training step. This iterative process is necessary because quantization itself is not differentiable~\citep{NEURIPS2019_f8e59f4b}, requiring special gradient accumulation methods. As a result, the high-precision weight matrices are always maintained throughout the whole training process, which makes traditional QAT inefficient and takes up a lot of extra memory footprints.  
For example, maintaining the weights of a 1B LLM alone would require 4GB of memory in the FP32 format, whereas maintaining ternary weights reduces this to 0.2GB. This requirement for memory footprints limits the accessibility of QAT techniques for researchers and organizations with limited computational resources.

In order to address these challenges, we explore Direct Quantized Training~(DQT), a modified QAT approach for language models that maintains only low-precision weights throughout the training process. DQT eliminates the reliance on STE by directly updating the quantized low-precision weights during backpropagation, as shown in Figure~\ref{fig_1} (b). Specifically, we use stochastic rounding~\citep{von1947numerical} to 
preserve the low-precision format of the weight matrices after backpropagation and minimize the information loss caused by using only low-bit weights. The DQT method keeps all weight matrices fixed at \textit{n}-bit precision~(INT\textit{n}) throughout the entire training process and eliminates the needs to quantize high-precision weight matrices at each training step. 
\textcolor{black}{More importantly, traditional QAT often introduces extra memory overhead due to the updates on high-precision weights, which limits its practicability in training general LLMs. In contrast, the light-weighted memory dependency of DQT enables quantization's application even for scenarios where computational resources are constrained.}


Experimental results on our LLaMA-structured models ranging from 130M, 320M and 1B parameters demonstrate that (1) DQT enables model convergence even when weight matrices are constrained to ternary values; (2) with 8-bit DQT, our models can achieve performance levels competitive with BitNet b1.58, showing the feasibility of the approach; (3) DQT models exhibit robust performance against GPU memory reductions, showing their light-weighted memory dependency; and (4) inference using only ternary weights in DQT remains effective, delivering performance comparable to BitNet. In addition, we conduct an in-depth analysis of stochastic rounding and find that it helps preserve critical update signals and contributes to training stability. We assume that DQT could provide new insights on addressing the computational challenges posed by traditional QAT.

\section{Related Works}
\label{related_works}
Efficient learning methods for LLMs have become a critical area of research~\citep{zhao-etal-2025-prompt}. Quantization for deep neural networks has a history spanning nearly ten years, with researchers initially compressing networks to reduce memory usage and computational load while maintaining accuracy~\citep{10.1007/978-3-319-46493-0_32, JMLR:v18:16-456}. Recent quantization methods for LLMs can be divided into two categories: Post-Training Quantization~(PTQ) and Quantization-Aware Training~(QAT).
\paragraph{Post-Training Quantization}
PTQ transforms high-precision parameters into low-bit ones after the parameters are already pre-trained. Some approaches utilize a small set of calibration data to accomplish this transformation while preserving the model's performance~\citep{10.5555/3524938.3525605, librecq, frantar2023gptq}. Others explore methods that eliminate the need for calibration data altogether~\citep{cai2020zeroq}. The challenge of PTQ lies in achieving a balance between compression efficiency and minimal accuracy degradation, often involving computational trade-offs that make it a non-trivial task for general-purpose applications. Moreover, the performance of PTQ consistently lags behind that of QAT, as there is a gap between the learned high-precision representations and the constrained bit width~\citep{chen2024efficientqatefficientquantizationawaretraining, liu-etal-2024-llmqat}.

\paragraph{Quantization-Aware Training}
To bridge the gap between high-precision parameter training and quantization, QAT incorporates the quantization of model parameters during the training process. 
One of the initial applications of QAT to LLMs comes from \citet{liu-etal-2024-llmqat}, who propose a data-free distillation-based method and quantize the model to 4 bits. \citet{xu2024onebit} further expand distillation-based methods to binary quantization by introducing limited trainable vectors.
More recently, BitNet~\citep{bitnet, b1.58} is proposed, achieving training from scratch QAT with weight values constrained to $\{-1,0,1\}$.  However, due to the non-differentiability of the quantization process, special gradient approximation methods like the straight-through estimator~(STE)~\citep{straight_through} are commonly employed during training. While effective, this approach often results in slower training processes and increased computational memory overhead since high-precision weights are always maintained during the training process. 
These inefficiencies become particularly pronounced when scaling QAT to larger models, limiting the practical training of QAT in real-world use cases. To address this, in this work, we explore the potential for more efficient and practical QAT methods that have significantly lower memory dependency compared to traditional QAT approaches.

\section{Method}
In this section, we first introduce stochastic rounding, the core idea of our proposed method that maintains weight matrices at low-precision during training. Then we move on to describe how stochastic rounding is applied in the modified training process for QAT. 
\subsection{Stochastic Rounding}
The idea of stochastic rounding~(SR) originates from \citet{von1947numerical}, and is initially used for reducing bias in numerical computations and recently for deep learning models~\citep{stochastic1}. It is a rounding technique that probabilistically rounds values to the nearest representable precision based on their distance from those values. Given a 
high precision value $x$, stochastic rounding can be defined as the following equation~\citep{stochastic1, pmlr-v202-markov23a, stochastic2}:
\begin{equation}
\label{sr}
    \mathrm{SR}(x) = \left\{
    \begin{aligned}
    & \lfloor x \rfloor, \ \mathrm{with} \  p = \lceil x \rceil - x  \\
    & \lceil x \rceil, \ \mathrm{otherwise} \\
    \end{aligned}
    \right.,
\end{equation}
where $p$ stands for the probability of turning $x$ to $\mathrm{floor}(x)$ or $\mathrm{ceil}(x)$. In this way, we can naturally quantize high-precision values into low-precision ones.

\subsection{Modified Training Process}
Next, we continue to explain the details of DQT. 
As shown in Figure~\ref{fig_1}~(b), we start training from low-precision weight matrices instead of high-precision ones. We achieve this through utilizing AbsMean Quantization following \citet{b1.58} on a randomly initiated weight matrix $W$.
The absolute mean value for $W$ can be represented as
\begin{equation}
\mathrm{AbsMean}(W) = \frac{1}{k}\sum_{i=1}^{k} \ \lvert w_i \rvert,
\end{equation}
where $w_i$ is the $i_{th}$ element in $W$. For $n$-bit quantization, $Q_n=-2^{n-1}$ and $Q_p=2^{n-1}-1$ are then determined to constrain the range of quantization, and the scaling factor $s$ can be defined as
\begin{equation}
s = \frac{Q_p}{\mathrm{AbsMean}(W)}.
\end{equation}
Finally, the quantized weights $\widetilde{W}$ can be expressed in the following equation:
\begin{equation}
\widetilde{W} = \mathrm{Clip}[\mathrm{Round}(W\cdot s), Q_n, Q_p] / s,
\end{equation}
where the $\mathrm{Clip}()$ function assures that all the values are in the range of $[Q_n, Q_p]$ and $\mathrm{Round}()$ returns the nearest integer. This allows us to constrain the quantized $\widetilde{W}$ to $n$-bits~(INT\textit{n}). While for the inputs and activations, we follow the settings introduced in BitNet~\citep{bitnet, b1.58} and quantize them into 8 bits. 

After computing the language modeling cross-entropy loss based on $\widetilde{W}$ and inputs in each training step, we first allow the optimizer to calculate $W'$, the dense weight matrix intended for updating. In traditional QAT, the straight-through estimator is applied, and $W'$ replaces the original high-precision $W$, subsequently undergoing the quantization process from Equation (2) to Equation (4) again in the next training step.  However, in our approach, we simplify this process by directly applying stochastic rounding on $W'$ using Equation~(\ref{sr}):
\begin{equation}
\widehat{W} = \mathrm{SR}(W'),
\end{equation}
to make sure it maintains $n$-bits~(INT\textit{n}) without requiring the retention of high-precision weights, thus getting rid of the straight-through estimator and skipping the process from Equation (2) to Equation (4). During our modified DQT, we directly replace $\widetilde{W}$ with $\widehat{W}$ and proceed to the next training step, ensuring that the weight matrices are always constrained to $n$-bits throughout training, which makes the biggest difference from traditional QAT.

\section{Experiments}
In this section, we begin by detailing our training dataset and implementation setup in Section~\ref{implementation}. Section~\ref{dense} presents a comparison between the training behavior of DQT models and other baselines. To demonstrate the robustness of DQT under reduced GPU memory environments, we evaluate its performance in FP32, BF16, and FP8 formats, as well as with memory-efficient optimizers, in Section~\ref{low_memory}. We then analyze the effect of varying bit widths in DQT in Section~\ref{bit-width}. Finally, Section~\ref{bitnet} reports evaluation results on various tasks.

\subsection{Implementation Setup}
\label{implementation}

\paragraph{Model architecture.} We follow the architecture of LLaMA2~\citep{touvron2023llama2} and initialize models with sizes ranging from 130M to 320M and 1B parameters. Detailed model specifications and configurations are provided in the Supplementary Material, Section 1.1. 
For the main experiments, we use AdamW~\citep{adamw} as the optimizer. For DQT models, we apply modified versions of the optimizers that incorporate stochastic rounding.

\paragraph{Baselines.} We choose two types of baselines for comparison with our DQT. The first is a reproduced full-precision~(FP32) model. We compare DQT to it to demonstrate the effectiveness of our approach with reduced bits.
The second baseline is the reproduced BitNet b1.58, a well-known QAT method that employs ternary weights at inference but relies on high-precision weights during training. 
\paragraph{Datasets} We use two types of datasets to pretrain the models. The first is the English Wikipedia dataset~(20231101.en)\footnote{\url{https://huggingface.co/datasets/wikimedia/wikipedia}}. For the 1B models, we further pre-train them using a larger dataset containing 10B tokens from the FineWeb dataset~\citep{penedo2024finewebdatasetsdecantingweb}\footnote{\url{https://huggingface.co/datasets/HuggingFaceFW/fineweb/viewer/sample-10BT}}. We split 1\% of the data as the corresponding development set. Models are trained for one epoch with a cosine scheduler applied and a 2000 step warm-up. Refer to Supplementary Material, Section 1.2 for details on how we process the training dataset.

\paragraph{Evaluation.} Besides reporting the training loss and perplexity results, we perform zero-shot evaluation on general language modeling tasks using the lm\_eval benchmark~\citep{lmeval-harness}. Specifically, we evaluate on WinoGrande~\citep{sakaguchi2019winogrande}, ARC~\citep{arcClark2018ThinkYH}, PIQA~\citep{piqaBisk2020} and SciQ~\citep{sciqWelbl2017CrowdsourcingMC}, which span a range of reasoning and question answering tasks across diverse domains.


\begin{figure*}[ht]
  \centering
  \includegraphics[width=0.48\textwidth]{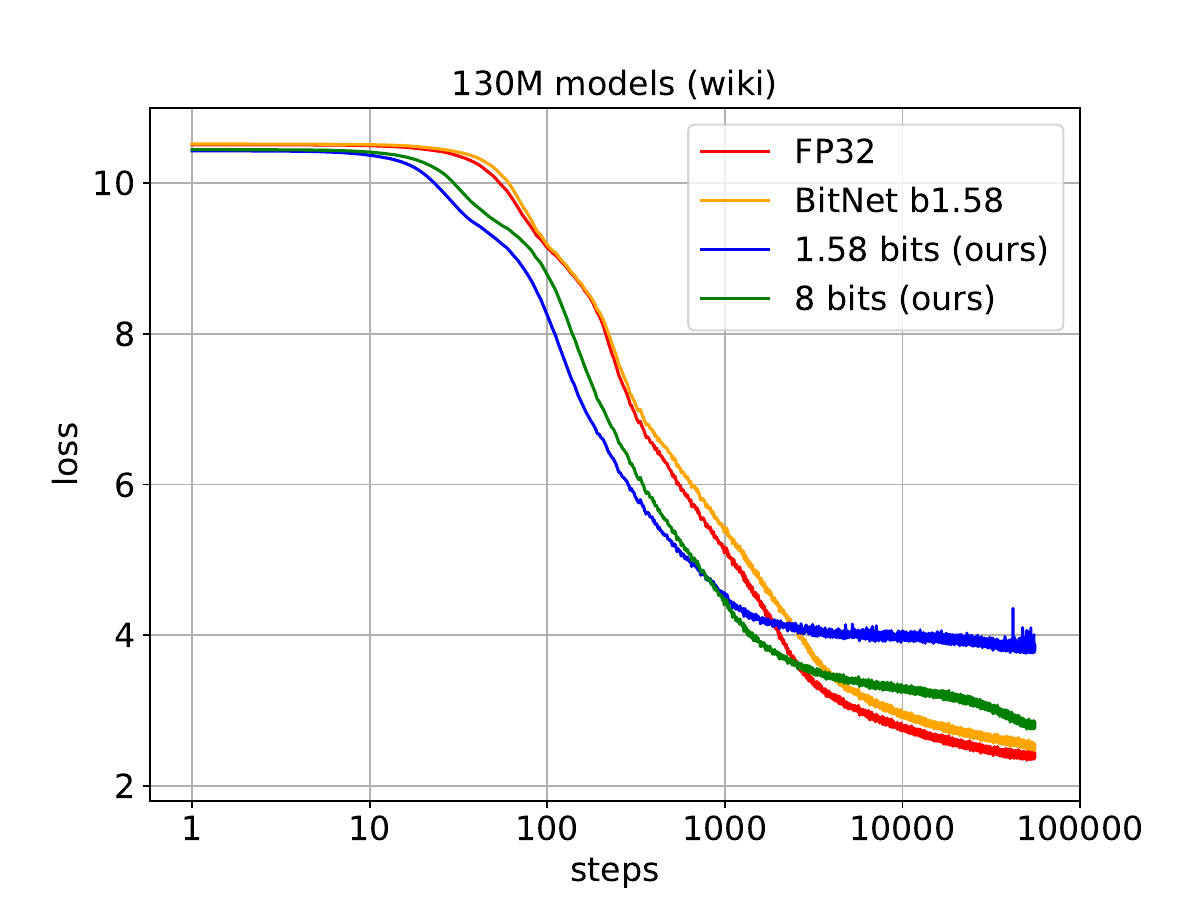}
  \includegraphics[width=0.48\textwidth]{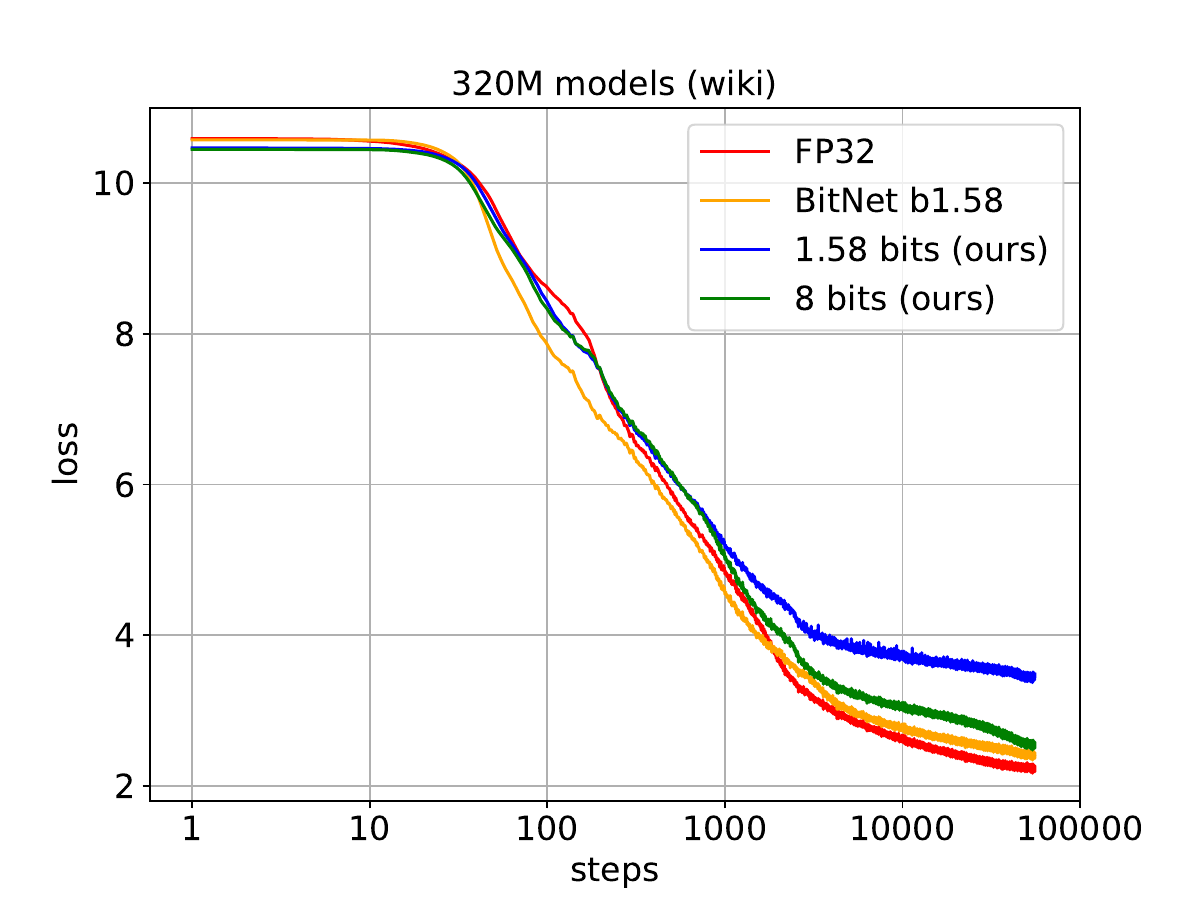}
  \includegraphics[width=0.48\textwidth]{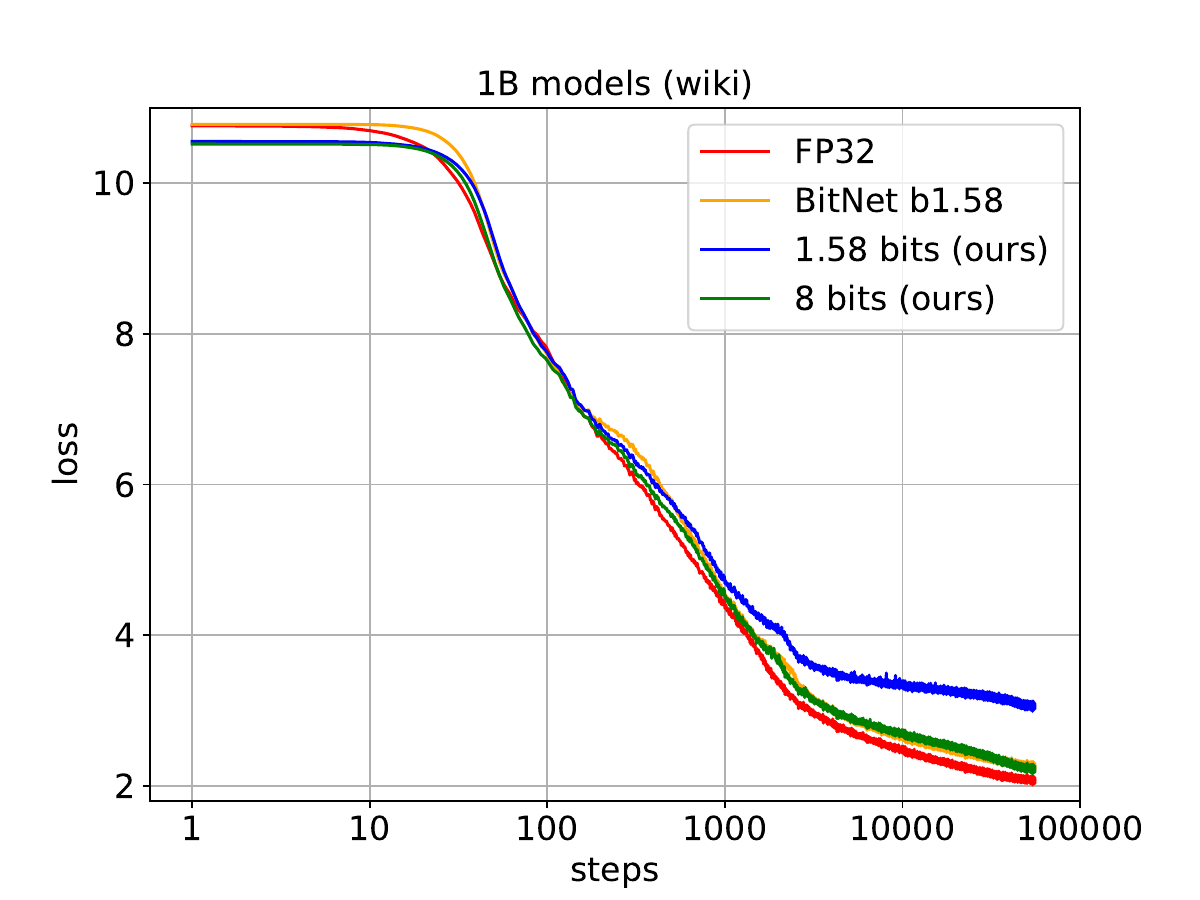}
  \includegraphics[width=0.48\textwidth]{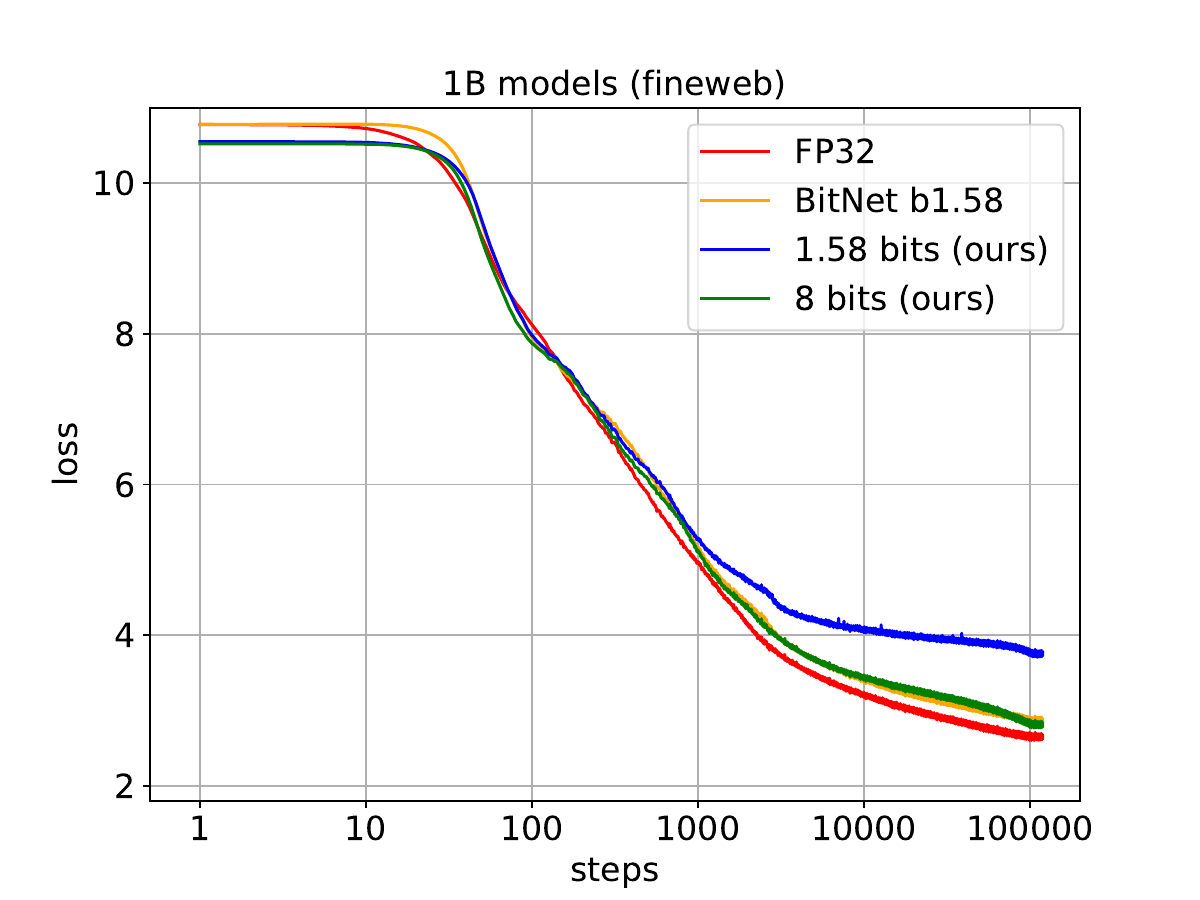}
  \caption{Comparison of our DQT and other baselines across different model sizes and training datasets. The horizontal axis represents the training steps while the vertical axis represents the training loss. As model size increases, the performance of our DQT models, especially DQT 8 bit, become more comparable to and even better than the reproduced BitNet b1.58.}
  \label{fig_2}
\end{figure*}
\paragraph{Hardware.} We pre-train the 130M models on 4 NVIDIA A100 80GB GPUs, the 330M models on 8 NVIDIA GH200 Grace Hopper Superchips, and the 1B models on 16 GH200 Superchips. To remain hardware-agnostic, we simulate our quantization approach under FP32, BF16, and FP8 environments. We use MS-AMP\footnote{\url{https://github.com/Azure/MS-AMP}} for FP8 experiments with optimization level set to MS-AMP O2.
While we recognize that FP8 is still not true ternary precision, it offers a practical compromise under current hardware constraints, as done in many prior works. 

\subsection{Main Results}
\label{dense}
We first present the training loss for DQT variants, our reproduced LLaMA~(FP32) and BitNet b1.58 in Figure~\ref{fig_2}, including different model sizes and training datasets under the FP32 environment. 
In all subfigures of Figure~\ref{fig_2}, the blue line represents the ternary implementation of DQT, where weight matrices are always constrained to $\{-1, 0, 1\}$. The green line denotes the 8-bit (INT8) implementation of DQT (in the format of FP32 due to simulation). The orange line corresponds to the reproduced BitNet b1.58, which utilizes high-precision information during training. Finally, the red line indicates the standard FP32 LLaMA implementation.

We first examine the blue lines: although there remains a performance gap compared to higher-precision models, the ternary DQT implementation still demonstrates the ability to converge. 
Across all model sizes and training datasets, we observe that standard FP32 models (red lines) consistently achieve the best performance, as they are not subject to any quantization. 
BitNet (orange lines) consistently delivers performance close to FP32, benefiting from high-precision updates during training.
Finally, for our DQT 8 bit implementation (green lines), the performance gap with BitNet b1.58 narrows as model size increases. Particularly, in the 1B models, DQT 8 bit models surpass BitNet b1.58, and this trend persists even when training on the larger FineWeb dataset. A clearer non-logarithmic comparison between the 1B DQT 8-bit model and BitNet b1.58 is provided in the Supplementary Material, Figure~1.

Generally speaking, as model size increases, all models show improved performance, reflecting the general benefits of scaling. However, the performance gains of our DQT models are especially pronounced. This suggests that DQT benefits more from scaling than other quantized approaches, narrowing the gap with other baselines and, in some cases, even surpassing them. Notably, since our DQT models do not rely on high-precision weights during training, even when 8-bit quantization is used, their memory requirements are significantly lower than those of BitNet, which still depends on high-precision updates. 
In the next section, we demonstrate DQT's low GPU memory dependency in detail.





\subsection{Low Memory Experiments}
\label{low_memory}
\begin{figure}[t]   
    \centering
    \includegraphics[width=0.7\textwidth]{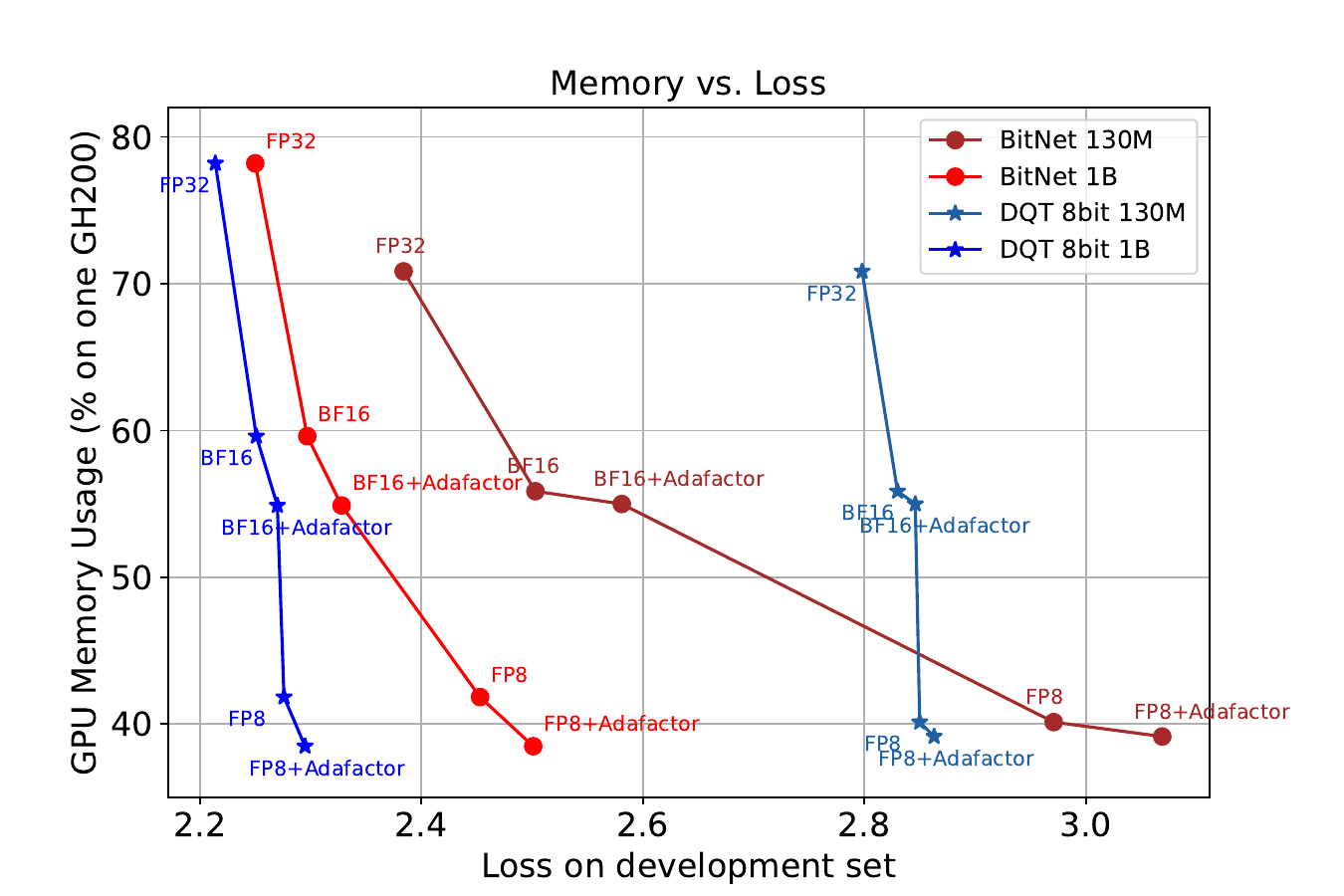}
    \caption{GPU memory usage versus loss on development set. While BitNet suffers significant performance degradation in low-precision formats, DQT demonstrates strong robustness with minimal loss increase.}
    \label{fig_mvsl}

\end{figure}
We additionally conduct two types of experiments to evaluate DQT's performance given reduced GPU memory, simulating resource-limited environments. First, we assess performance in low-precision formats using BF16 and FP8 \footnote{In our notation,``n-bit'' refers to the quantization level of the model weights (INT-n), while the labels such as FP32, BF16, or FP8 refer to the precision of the environment. Note that any ``n-bit'' can be simulated in these precisions.}. Second, we examine the effect of memory-efficient optimizers, as standard AdamW maintains two high-precision states (momentum and variance) for each parameter, contributing significantly to memory usage.
To demonstrate DQT’s low reliance on high-precision information, we specifically choose Adafactor~\citep{shazeer2018adafactor}, which eliminates the need for fully storing these additional states. Note that we conduct experiments with Adafactor in BF16 and FP8 formats, as its benefits are most pronounced in low-precision environments. 

Figure~\ref{fig_mvsl} presents the GPU memory usage and corresponding development set loss at the end of training for 130M and 1B models trained on the Wikipedia dataset. The y-axis indicates the percentage of GPU memory used on a single GH200 GPU under our experimental settings. We explain how we acquire the percentage in Supplementary Material, Section 1.4. As shown in the figure, BitNet models experience clear performance degradation as GPU memory usage decreases in BF16 and FP8 settings, for both 130M and 1B scales. This is expected, since BitNet still relies on high-precision information for effective weight updates. In contrast, our DQT 8-bit models maintain robust performance under reduced precision. For both 130M and 1B models, the performance drop is less than 0.1 in loss, demonstrating DQT’s resilience to lower GPU memory usage and its minimal dependence on high-precision information. This robustness persists even when applying memory-efficient optimizers such as Adafactor in both BF16 and FP8 settings.
These results suggest that DQT models can be effectively trained even in memory-constrained environments, showing consistent performance across both low-precision formats and lightweight optimization methods. We believe that DQT can be further extended to even lower-precision settings while maintaining its effectiveness.

\subsection{Impact of the Bit Width in DQT}
\label{bit-width}
\begin{figure}[t]   
    \centering
    \includegraphics[width=0.48\textwidth]{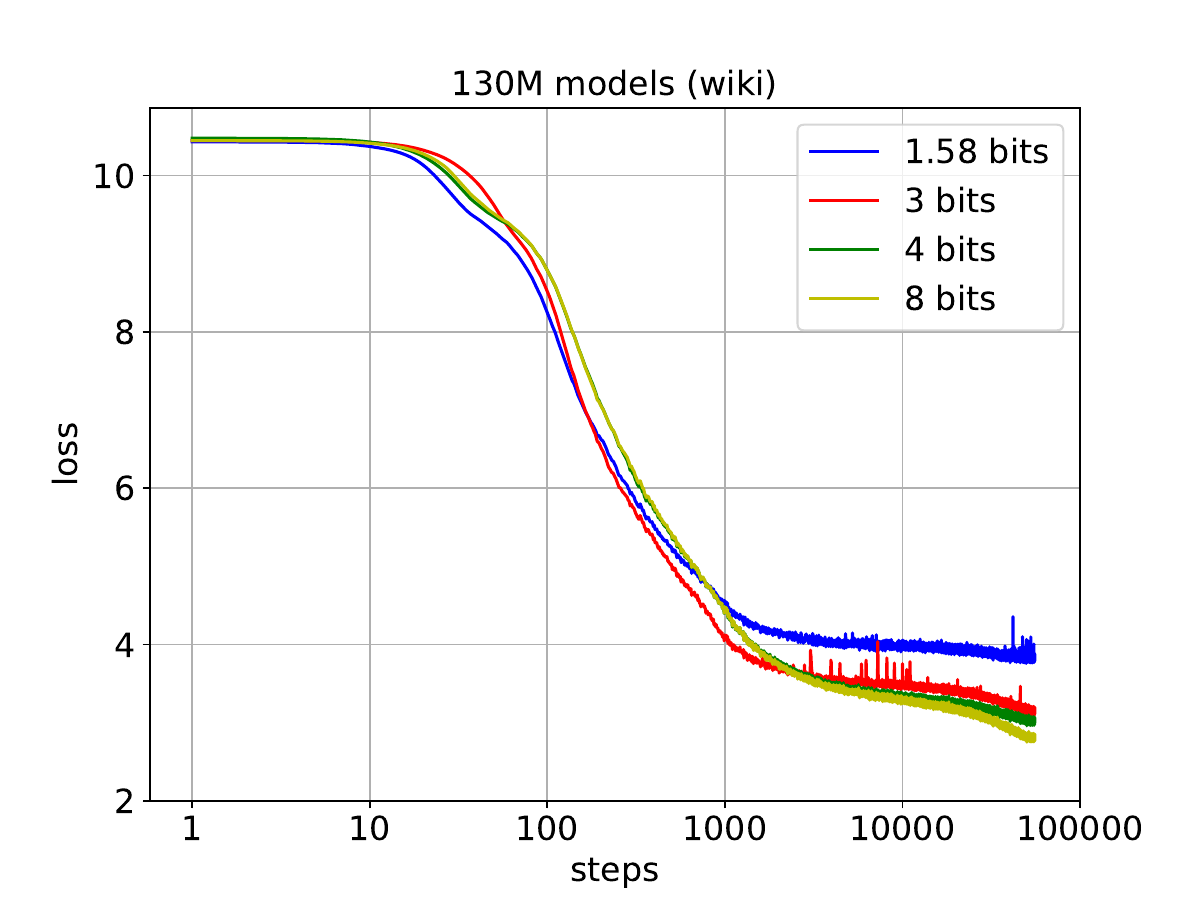}
    \includegraphics[width=0.48\textwidth]{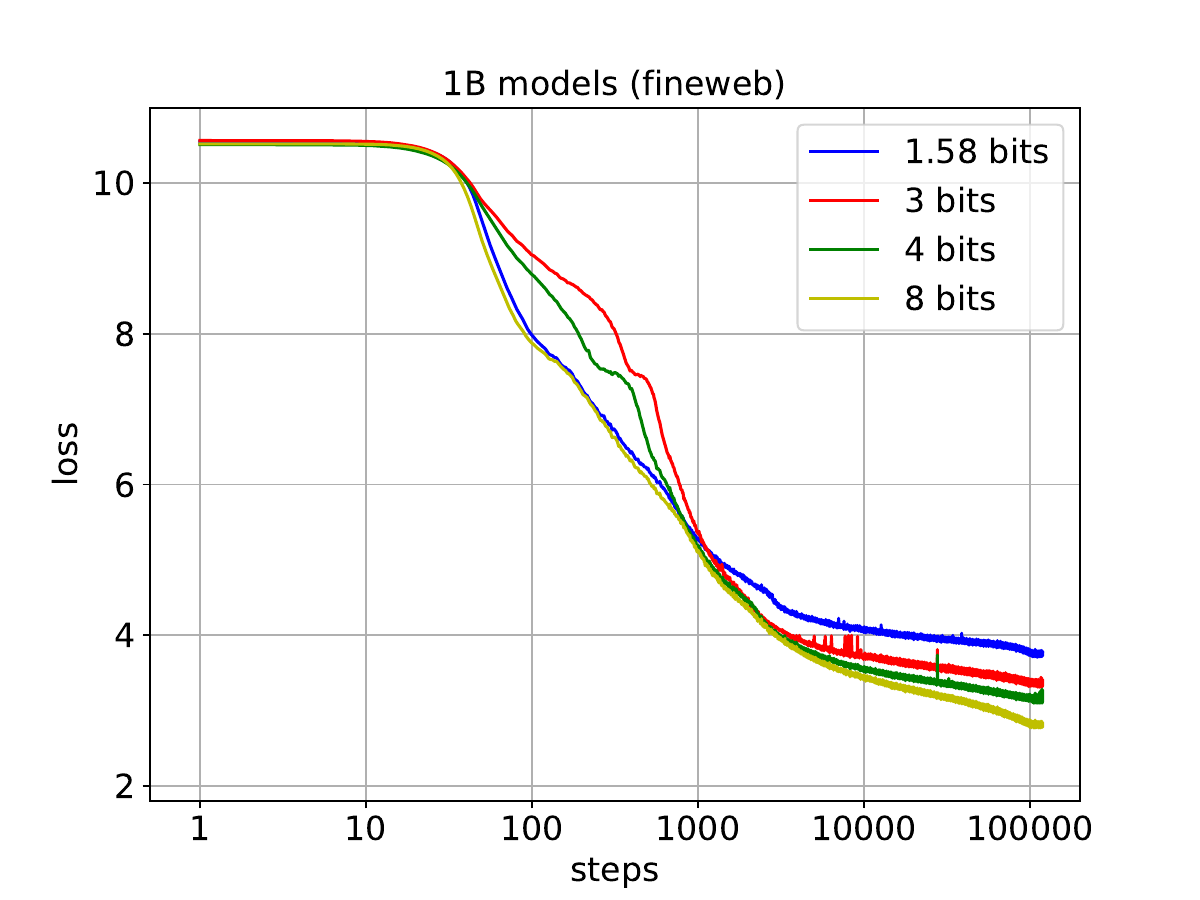}
    \caption{Comparison of bit widths in DQT. Higher $n$-bit results in better performance. }
    \label{fig_3}

\end{figure}
Next, we proceed to examine the impact of varying the number of bits in DQT. Specifically, we experiment with $n$ in \{1.58, 3, 4, 8\} on 130M models trained with the Wikipedia dataset and 1B models trained with the FineWeb dataset under FP32. The corresponding results are presented in Figure~\ref{fig_3}. We select the optimal learning rate for each model.

From Figure~\ref{fig_3}, we can first observe a clear trend: as the number of bits used in DQT increases, the model's performance improves consistently for both 130M and 1B models. Notably, for the relatively lower bits DQT implementations (1.58-bit and 3-bit), we can observe some outliers in the blue line and red line, indicating the difficulty of training low-bit models. In contrast, the higher-bit implementations exhibit more stability throughout the training process.

\subsection{Evaluation Results}
\label{bitnet}
\begin{table*}[t]
\centering
\resizebox{0.9\textwidth}{!}{
\begin{tabular}{ccccccc}
\hline
Models                          & Wikitext2($\downarrow$)           & WinoGrande ($\uparrow$)           & ARC (easy) ($\uparrow$)            & ARC (challenge) ($\uparrow$)       & PIQA  ($\uparrow$)                 & SciQ ($\uparrow$)                  \\ \hline \hline
\multicolumn{7}{c}{\textit{1B models (wiki)}}                                                                                                                                                 \\ \hline 
FP32                           &     27.03                  & 49.01                   &      39.56             &      23.29            &   56.20                 &       70.70                \\
BitNet b1.58                    &     34.90                  & 51.38                  & 36.74                & 22.95                    & 54.68                    &    69.40                   \\
DQT 8 bit & 30.94 & 49.96                     & 37.54                     & 23.55                     & 56.47                     & 69.50 \\ 
DQT 8 bit~(ternary Inf.) & 35.51 & 51.30                     & 36.36                     & 24.23                     & 54.19                    & 68.90 \\
\hline
\multicolumn{7}{c}{\textit{1B models (fineweb)}}                                                                                                                                                 \\ \hline 
FP32                            &     19.99               & 52.33                 &      48.06            &      25.00           &   67.90               &      80.20                \\
BitNet b1.58                &    28.20                 & 51.22                 & 44.36                 & 22.53                   & 65.40                    &   75.40                    \\
DQT 8 bit & 25.43 & 51.97                     & 45.12                     & 23.63                    & 66.38                    & 75.90 \\ 
DQT 8 bit~(ternary Inf.) & 27.32 & 51.70                     & 45.62                     & 22.87                    & 65.51                     & 75.70 \\
\hline \hline
\end{tabular}}
\caption{Evaluation results on different tasks. Except for WikiText-2, where perplexity is reported, we report the accuracy metric for the remaining tasks.}
\label{main_table}
\end{table*}
In this section, we conduct evaluation on the WikiText-2 test set~\citep{merity2016pointersentinelmixturemodels} and several other zero-shot tasks. We focus on 1B parameter models to best demonstrate the capability of our approach. The results are summarized in Table~\ref{main_table}. For consistency, we set the sequence length to 512 across all tasks, matching the configuration used during training. In the table, `ternary Inf.' refers to DQT models trained with 8-bit weights but evaluated using ternary weights. Details on how ternary inference is implemented for DQT models are provided in Supplementary Material, Section 3.
As shown in Table~\ref{main_table}, FP32 models achieve the highest overall performance across tasks, followed closely by our DQT 8-bit variants. Notably, except for the WinoGrande task using 1B models trained on the Wikipedia dataset, our DQT 8 bit models outperform BitNet b1.58 across all other benchmarks. These results highlight that DQT 8 bit models more closely approximate the performance of FP32 models compared to BitNet.
Moreover, when ternary inference is applied, the performance slightly decreases compared to 8-bit inference but remains on par with BitNet, demonstrating the robustness of our approach with ternary inference and its flexibility in deployment.


\section{Analysis}
The use of stochastic rounding in DQT can be viewed as a convergence-guaranteed optimization scheme. This follows from the fact that stochastic rounding introduces zero-mean noise with bounded variance, a property well studied in stochastic optimization theory~\citep{stochastic1, JMLR:v18:16-456, bottou2018optimizationmethodslargescalemachine, ajalloeian2021convergencesgdbiasedgradients}. As a result, the updates in DQT maintain convergence guarantees despite operating in a quantized space. For completeness, we provide a proof sketch of this guarantee in Supplementary Material, Section 4. In this section, we primarily focus on the empirical analysis, demonstrating how these theoretical properties manifest in practice.

While stochastic rounding offers advantages in keeping low-precision weights,
it may bring some problems. For example, small weight updates may be applied with low probability (e.g., a value like $-1+0.02$ has only a 2\% chance of being rounded to 0). If such rare updates do occur, we are particularly interested in understanding their impact on the overall training dynamics. In this section, we provide a detailed analysis of the effects and implications of using stochastic rounding.

\subsection{The Role of Stochastic Rounding in DQT}
\label{5.1}
\begin{figure}[t]   
    \centering
    \includegraphics[width=0.48\textwidth]{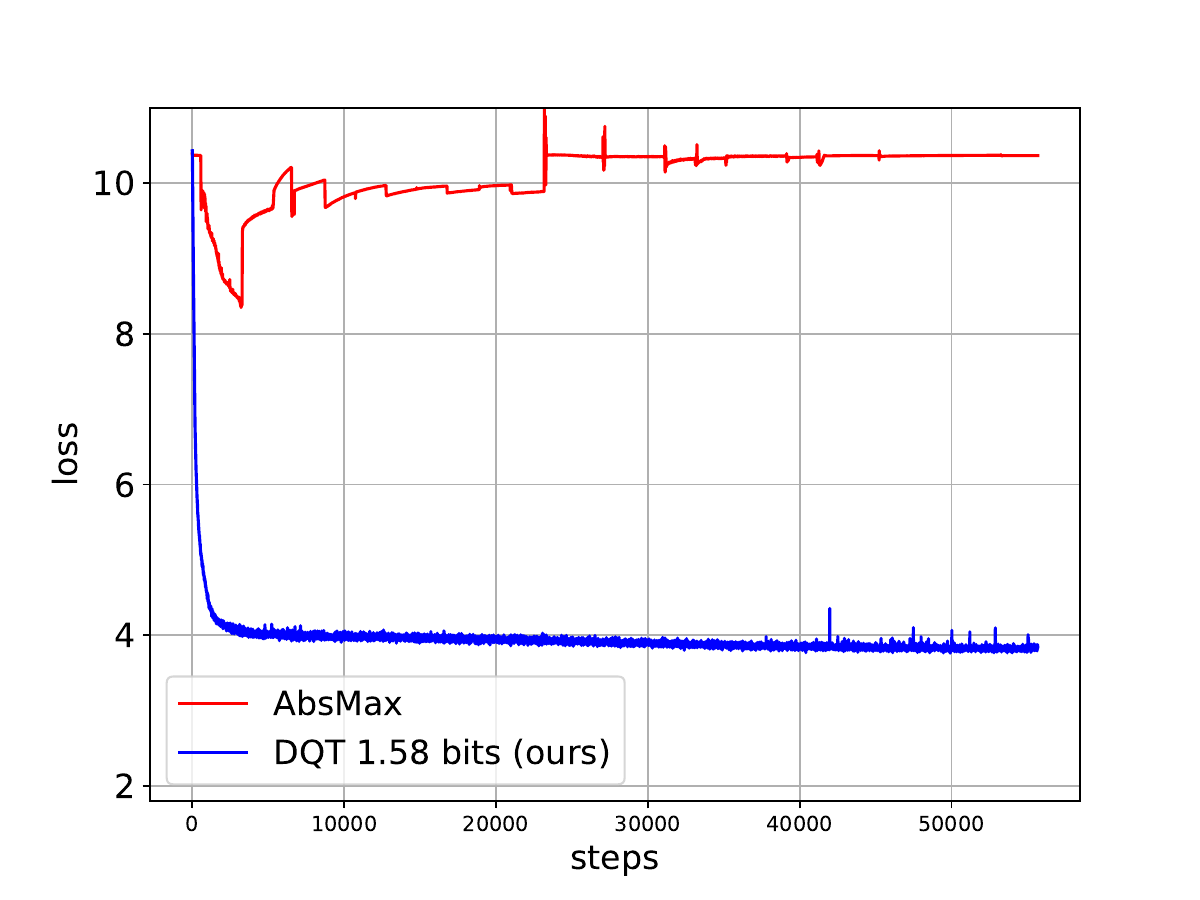}
    \includegraphics[width=0.48\textwidth]{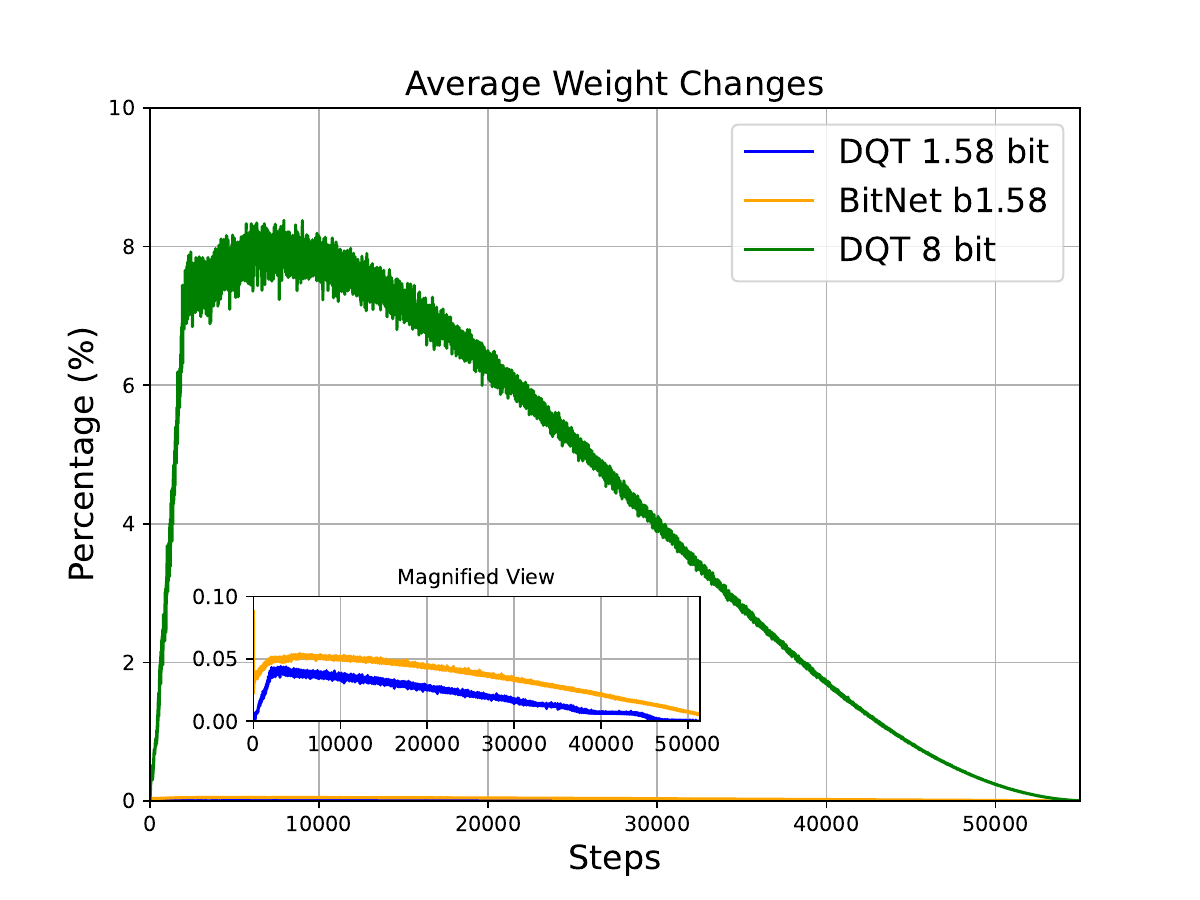}
    \caption{Left: Comparison between DQT 1.58 bits and a variant using absmax quantization for weight updates under the same learning rate. Right: Percentage of updated weights after each training step.}
    \label{fig_vs}

\end{figure}

Firstly, we provide empirical evidence for the critical role of stochastic rounding in preserving gradient information during training with low-precision weights. Particularly, we compare 130M ternary DQT with a variant that simply uses absmax quantization on the updated weight matrices and maintains the weights in ternary format without stochastic rounding. As shown in the left part of Figure~\ref{fig_vs}, the latter fails to converge, despite operating under the same bit budget. This outcome is expected, as absmax quantization can easily ignore small updates on the weights. In contrast, stochastic rounding not only performs quantization but also facilitates the accumulation of fine-grained updates, enabling effective training even in extremely low-bit regimes.

\subsection{Quantifying Weight Update Frequency in DQT}
As discussed in Section~\ref{5.1}, stochastic rounding facilitates training by allowing even small weight updates to take effect. 
To better understand its impact, we examine how frequently weights are updated during training. Specifically, we analyze 130M-size models to measure the percentage of quantized weights that change after each training step.

We quantify the weight update frequency in the right part of Figure~\ref{fig_vs} for three variants: DQT 1.58 bit, BitNet b1.58 and DQT 8 bit under the same learning rate and batch size. Note that we show the average percentages of all weight matrices in the model. For BitNet, the peak weight update rate is approximately 0.05\% (observed at step 2000, the end of the warm-up phase), meaning only 0.05\% of quantized ternary weights change after a single step. Ternary DQT exhibits a similar update rate of around 0.04\%, indicating minimal difference between the two in terms of update frequency. In contrast, the 8-bit DQT variant, with weight values ranging from $-128$ to 127, shows a significantly higher update frequency, reaching up to 8\%.

\subsection{Impact of Small Weight Updates in DQT Training}
\begin{figure}[t]   
    \centering
    \includegraphics[width=0.6\textwidth]{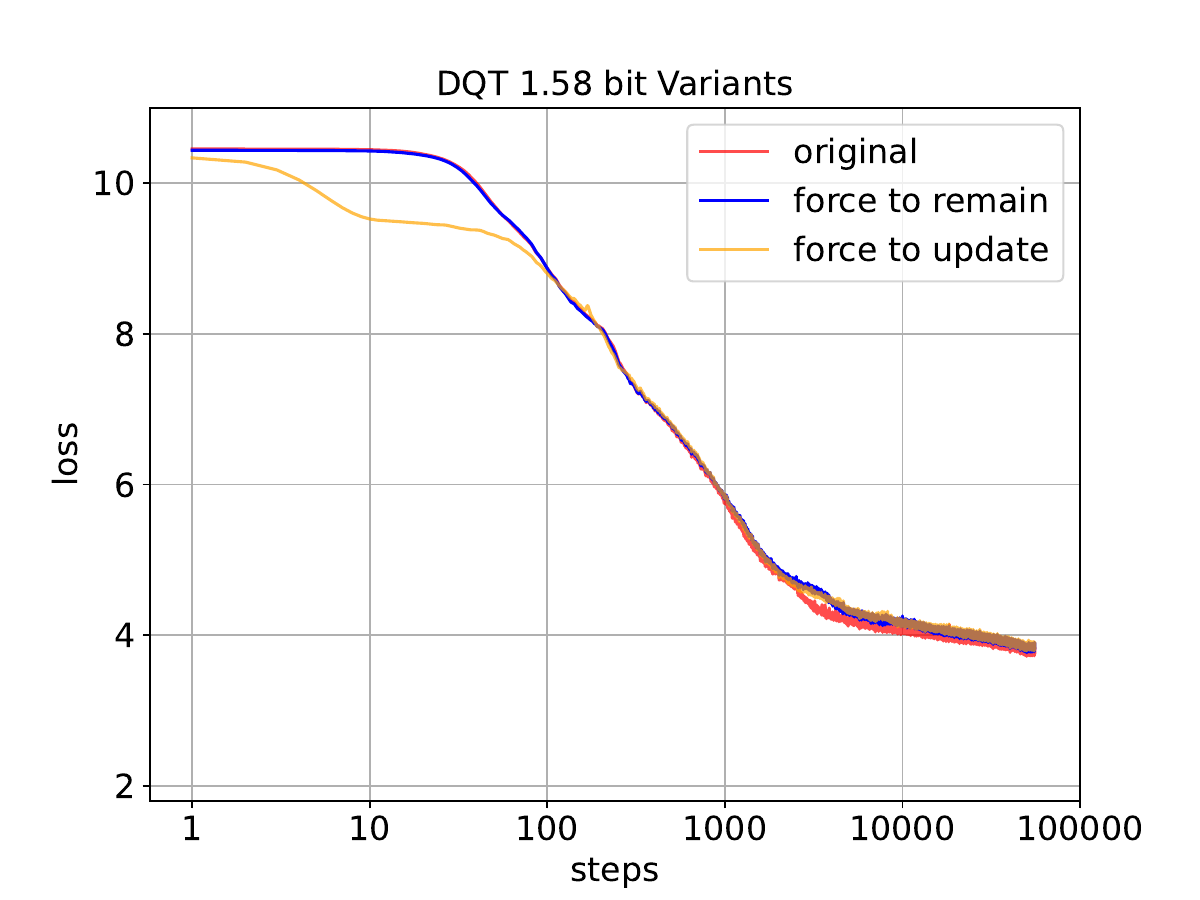}
    \caption{Force to remain refers to the variant that ignores the smallest 20\% of weight updates, while force to update enforces changes in this bottom 20\%. All variants use the same learning rate.}
    \label{fig_change_unchange}

\end{figure}
Finally, to investigate the impact of small weight updates during training, we rank the absolute value of weight updates at each step and selectively intervene in the lowest 20\%. For these bottom 20\% updates, we apply one of two interventions: either suppress the update by retaining the original ternary value, or enforce a change by rounding it to a different quantized value, even if the update is small. For example, if an update of +0.02 from $-1+0.02$ falls within the smallest 20\%, we either keep it at $-1$ (suppress) or round it to 0 (enforce), depending on the experimental condition. Figure~\ref{fig_change_unchange} shows the training loss for these variants, where \textit{force to remain} refers to suppressing the bottom 20\% of updates, and \textit{force to update} corresponds to enforcing them to change. From Figure~\ref{fig_change_unchange}, we observe that the original implementation of DQT (1.58-bit) achieves the best performance. Ignoring the smallest 20\% of updates has minimal impact, while enforcing these small updates to take effect slightly accelerates convergence. However, the final losses at the end of training are similar across all variants. This may be due to the fact that small updates contribute less to training overall, and even when such updates do occur, stochastic rounding may re-adjust them toward the optimal value in the next training steps.

\section{Conclusion}
In this work, we explore Direct Quantized Training, a modified QAT method that directly updates low-bit weight matrices without relying on high-precision weights during training. This design enables DQT to operate effectively even under constrained GPU memory settings. Experimental results across different sizes of models demonstrate that DQT enables training without updating on high-precision weights, which are required for straight-through estimation. Moreover, when using 8-bit weights, DQT achieves performance comparable to both FP32 models and BitNet b1.58 during training and inference, demonstrating its effectiveness and practicality for efficient model training.

\bibliography{acml25}

\end{document}